\pdfoutput=1

\documentclass[11pt]{article}

\usepackage[preprint]{acl}

\usepackage{times}
\usepackage{latexsym}

\usepackage[T1]{fontenc}

\usepackage[utf8]{inputenc}

\usepackage{microtype}

\usepackage{inconsolata}

\usepackage{graphicx}
\usepackage{hyperref}
\usepackage{comment}

\usepackage{caption}
\usepackage{subcaption}
\usepackage[export]{adjustbox}
\usepackage{booktabs}
\usepackage{makecell}
\usepackage{graphicx}
\usepackage{tabularx}
\usepackage{multirow}
\usepackage{amsmath}
\usepackage{amssymb}
\usepackage{stmaryrd}
\usepackage{enumitem}

\title{Evaluating Adjective-Noun Compositionality in LLMs: \\ Functional vs Representational Perspectives}

\author{Ruchira Dhar$^{\dagger}$ \ \ 
        Qiwei Peng$^{\dagger}$ \ \ 
        \textbf{Anders S{\o}gaard}$^{\dagger}$ \\
        $^{\dagger}$Department of Computer Science, University of Copenhagen \\ 
        }

\begin{document}
\maketitle
\begin{abstract}

Compositionality is considered central to language abilities. As performant language systems, how do large language models (LLMs) do on compositional tasks? We evaluate adjective–noun compositionality in LLMs using two complementary setups: prompt-based functional assessment and a representational analysis of internal model states. Our results reveal a striking divergence between task performance and internal states. While LLMs reliably develop compositional representations, they fail to translate consistently into functional task success across model variants. Consequently, we highlight the importance of contrastive evaluation for obtaining a more complete understanding of model capabilities. 

\end{abstract}

\section{Introduction}

Compositionality ---the ability of systems to generate complex meanings from simpler parts---  is often considered a key factor in human language processing \cite{ smolensky1987connectionist,fodor1988connectionism, partee1995lexical, townsend2018compositionality,donatelli2023compositionality}. Today, large language models (LLMs) are considered highly performant natural language systems \cite{10.1145/3605943}. This has led to questions on the compositionality of such models \cite{saba2023stochastic,mahowald2024dissociating, mccurdy-etal-2024-toward}: if compositionality is key to language, then are LLMs compositional as well?\footnote{Non-compositional LLMs have been argued to at least be conceivable \cite{Block1981-BLOPAB,bender2021dangers}.}

Early empirical work investigated compositionality in neural models with tasks like SCAN \cite{lake2018generalization}, COGS \cite{kim-linzen-2020-cogs}, and PCFG \cite{hupkes2020compositionality}. However, these paradigms are a poor fit for modern pretrained LLMs: they rely on controlled train–test splits designed for models trained from scratch, and many use synthetic or restricted fragments of English that are likely already represented in large-scale pretraining corpora. More recently, compositionality has been evaluated in LLMs through functional task-based assessments \cite{pavlick-callison-burch-2016-babies,bertolini2022testing,sun2024evaluating,xu2024large,zhang2024can, dziri2024faith, ma2024examination}, alongside methods aimed at improving performance on such tasks \cite{shao2023compositional,drozdov2023compositional,press-etal-2023-measuring,lu2024chameleon,chen2024skillsincontext, li2024understanding}. In parallel, another line of work probes internal representations to identify compositional structure and layer-wise mechanisms \cite{hewitt-manning-2019-structural, murty2023characterizing, kumon-yanaka-2025-analyzing,peng2025understanding, guo2025quantifying}. Yet these functional and representational approaches are rarely evaluated side by side on the same models and tasks, making it difficult to determine whether they provide converging or divergent evidence about compositionality in LLMs.

Recent work suggests that external task performance and internal representations may diverge in LLMs \cite{lampinen2024learned, orgad2025llms} --- but does such divergence also apply to compositional abilities? Do functional task evaluations and representational analyses provide converging evidence of compositionality in LLMs? In this work, we address this gap by directly contrasting functional and representational evaluation within a unified experimental framework. We focus on adjective–noun compositionality and evaluate LLMs across three inference-based tasks capturing substitutivity, systematicity, and overgeneralization \cite{hupkes2020compositionality, hupkes2023taxonomy}. For each model family, we measure task performance under scaling and instruction tuning and conduct a layer-wise analysis to assess compositional structure in internal representations. By comparing behavioral trends with representational signals across model variants, we seek to test whether the two paradigms yield converging empirical conclusions about compositionality in LLMs.

\begin{table*}[ht!]
\centering
\renewcommand{\arraystretch}{1.2}
\setlength{\tabcolsep}{6pt}  
\resizebox{1\textwidth}{!}{%
\begin{tabular}{l *{12}{c}}
\toprule
\textbf{} &
\multicolumn{3}{c}{\textbf{LLaMA2}} & 
\multicolumn{3}{c}{\textbf{CodeLLaMA}} & 
\multicolumn{3}{c}{\textbf{Qwen2.5-Coder}} & 
\multicolumn{3}{c}{\textbf{Gemma2}} \\
\cmidrule(lr){2-4} \cmidrule(lr){5-7} \cmidrule(lr){8-10} \cmidrule(lr){11-13}
\textbf{Dataset} & \textbf{7B} & \textbf{7B-it} & \textbf{13B} & \textbf{7B} & \textbf{7B-it} & \textbf{13B} & \textbf{7B} & \textbf{7B-it} & \textbf{14B} & \textbf{2B} & \textbf{2B-it} & \textbf{9B} \\
\midrule
\textbf{AddOne (F1)}     & 77.97 & 18.01 & 78.49 & 78.49 & 75.36 & 54.92 & 77.58 & 20.87 & 12.62 & 78.49 & 29.61 & 59.76 \\
\textbf{PLANE (F1)}      & 36.70 & 33.73 & 35.35 & 33.33 & 42.67 & 33.33 & 33.33 & 34.57 & 38.14 & 33.33 & 33.33 & 36.84 \\
\textbf{CompComb (Acc)}  & 66.67 & 72.59 & 45.93 & 95.56 & 71.11 & 93.33 & 77.04 & 73.34 & 71.11 & 77.04 & 73.33 & 45.19 \\
\bottomrule
\end{tabular}%
}
\caption{The results of all models on Addone, PLANE, and COMPCOMB datasets.}
\label{tab:main_result}
\end{table*}

\section{Experimental Setup}
\label{sec:expsetup}

To study compositionality, we focus on the domain of adjective–noun (AN) phrases. From a linguistic perspective, AN constructions are among the most frequent and semantically transparent instances of phrasal composition, making them a classic testbed for compositional semantics \citep{partee1995lexical, kamp1995prototype, guevara2010regression, hartung2017learning, bertolini-etal-2021-representing}.  Crucially, meaning composition in this domain can be operationalized through inference: if the composed phrase preserves or alters entailment relations with its parts, the model must reason about semantic structure rather than rely on lexical associations \cite{lebowitz1983generalization, geiger-etal-2019-posing, donatelli2023compositionality}. This inference-based framing has been central in both formal semantics \citep{partee1995lexical} and recent NLP evaluations of compositionality \citep{pavlick-callison-burch-2016-babies,bertolini2022testing}.

\subsection{Task Choice}

Compositionality in natural language can manifest in many forms, such as substitutivity, systematicity, and overgeneralization \cite{hupkes2020compositionality}. We adopt one adjective-noun (AN) based task on meaning inference for each of these aspects: 

\paragraph{Substitutivity.} Substitutivity requires that substituting a semantically compatible modifier preserves entailment  \cite{pickel2004compositionality, werning2012oxford, hupkes2020compositionality, li2024compositional}. Let $S(n)$ denote a sentence containing noun $n$, and $S(an)$ its adjective-modified form. 
We define a binary entailment function:

\begin{equation}
E\big(S(an), S(n)\big) =
\begin{cases}
1 & \text{if } \llbracket S(an) \rrbracket \models \llbracket S(n) \rrbracket \\
0 & \text{otherwise.}
\end{cases}
\end{equation}

For example, entailment holds in 
\textit{The runner set a new record} $\models$ \textit{The runner set a record}, 
since the adjective restricts but does not alter the core meaning. 
In contrast, entailment does not hold in 
\textit{The alleged thief was arrested} $\not\models$ \textit{The thief was arrested}, 
as the adjective modifies the truth conditions of the noun phrase. 
We use the \textbf{AddOne} dataset \cite{pavlick-callison-burch-2016-babies} to evaluate substitutivity.

\paragraph{Systematicity.}
Systematicity requires recombining known semantic relations to infer a novel one \cite{cummins1996representations, symons2014systematicity, hupkes2020compositionality}. 
Given two base entailments---$an \models n$ and $an \models h$ (with $h$ a hypernym of $n$)---and must decide the composed entailment $an \models ah$, testing whether the adjective can be systematically carried across the noun--hypernym relation. We define a binary entailment function over the target inference:
\begin{equation}
E(an,ah) =
\begin{cases}
1 & \text{if } \llbracket an \rrbracket \models \llbracket ah \rrbracket \\
0 & \text{otherwise.}
\end{cases}
\end{equation}

We then consider three inference types: 
IT1: $E(an,n)$, 
IT2: $E(an,h)$, 
and IT3: $E(an,ah)$.

For some adjectives, all inference types hold:
\textit{red car} $\models$ \textit{car}, 
\textit{red car} $\models$ \textit{vehicle}, and 
\textit{red car} $\models$ \textit{red vehicle}. While for others, IT1 and IT2 hold but IT3 fails:
\textit{small elephant} $\models$ \textit{elephant}, 
\textit{small elephant} $\models$ \textit{animal}, but 
\textit{small elephant} $\not\models$ \textit{small animal}. 
 
We use the \textbf{PLANE} dataset \cite{bertolini2022testing} to evaluate systematic compositional inference.

\paragraph{Overgeneralization.} Overgeneralization occurs when compositional rules are incorrectly extended to expressions whose meanings do not transparently derive from their parts \cite{macken2014representation, hupkes2020compositionality, csordas2023systematic}.  In adjective--noun phrases, the modifier typically preserves type membership (e.g., $an$ denotes a subset of $n$). 
However, exocentric compounds may contain the same noun string while failing to denote a subtype of that noun. This task tests whether models avoid inferring type membership based solely on surface form overlap. Given a triple $(n, an, c)$, where $an$ is compositional and $c$ is an exocentric compound containing $n$, we define a binary type inference function:

\begin{equation}
E(x,n) =
\begin{cases}
1 & \text{if } \llbracket x \rrbracket \subseteq \llbracket n \rrbracket \\
0 & \text{otherwise.}
\end{cases}
\end{equation}

Compositional behavior predicts:
$E(an,n)=1$ and $E(c,n)=0$. For example,\textit{trenchcoat} $\models$ \textit{coat}, 
whereas 
\textit{turncoat} $\not\models \textit{coat}$. We introduce the \textbf{COMPCOMB} dataset to evaluate robustness against such overgeneralization. For more details on the three aspects and how the tasks represent them, refer to \autoref{sec:appendix} \footnote{We will release all code and task data upon publication.}. 

\subsection{Functional Evaluation} 
\label{sec:compab}

We describe the first of our two evaluation setups, which examines compositionality through task-level performance. Here, models are evaluated based on their behavioral outputs on the three benchmarks introduced above. We examine how compositional task performance varies under two factors commonly associated with improved LLM capability: \textbf{model scaling}, where performance tends to increase with parameter count \cite{kaplan2020scaling, hoffmann2022training, biderman2023pythia, chowdhery2023palm}, and \textbf{instruction tuning}, which has been shown to enhance generalization and alignment \cite{wei2022finetuned,ouyang2022training,longpre2023flan, chung2024scaling, sun2024evaluating}. 
If compositional behavior tracks broader capability improvements, systematic trends should be observable across scaling and instruction-tuned variants.

\paragraph{Methodology.}
We evaluate four model families—\verb|LLaMA-2|, \verb|CodeLlama|, \verb|Qwen2.5-Coder|, and \verb|Gemma2|—each with three variants: a base model, an instruction-tuned model, and a larger scaled model (details in \autoref{appendix:b}).  For AddOne and PLANE, we use two evaluation setup variations to mitigate effects of prompt sensitivity \cite{lu2024prompts, sclar2024quantifying} and the reported results correspond to average accuracy across these variants. For the prompting setup, we employ accuracy and for log probability setup, given two candidate completions (entailment vs.\ non-entailment), the model’s prediction corresponds to the completion with higher conditional probability. For COMPCOMB, we employ three prompting configurations designed to test type inference robustness (details in \autoref{appendix:c}). 

\begin{figure}
    \centering
    \includegraphics[width=1\columnwidth]{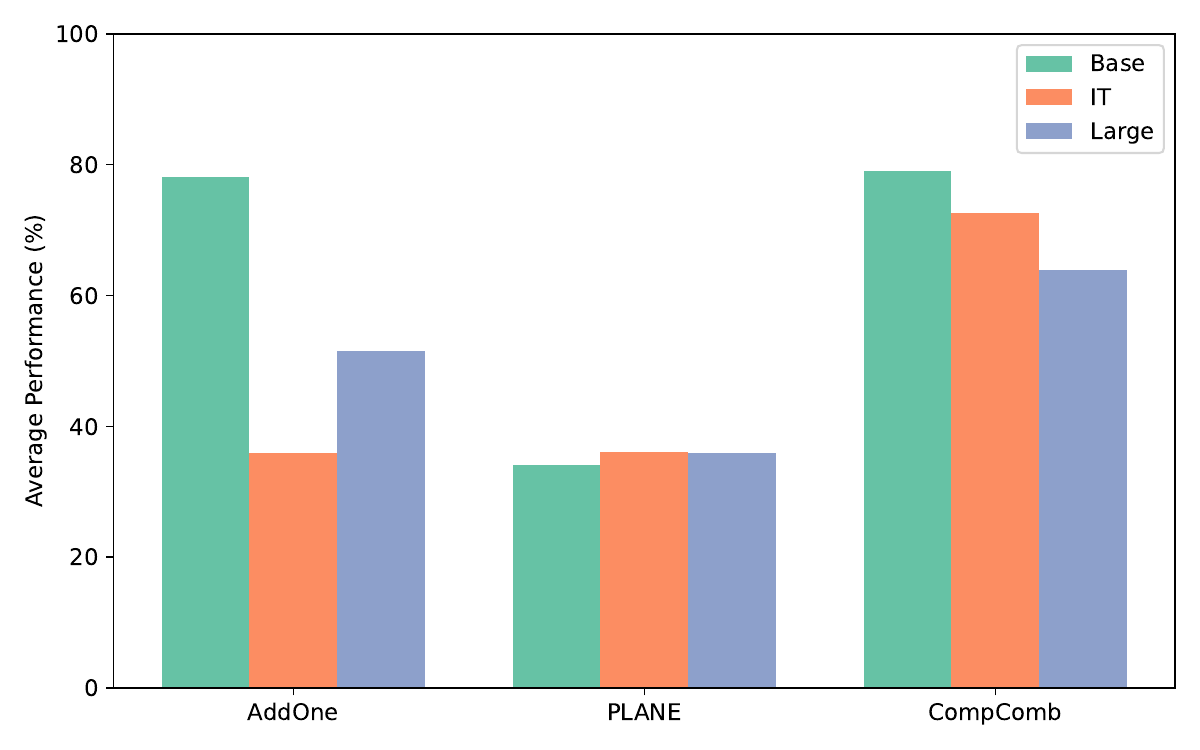}
    \caption{The average performance across different model category (Base, Instruction Tuning, and Large model size) on three tasks. We report the weighted F1 score on AddOne and PLANE, and Accuracy on COMPCOMB.}
    \label{fig:average_performance_across_category}
\end{figure}

\paragraph{Results.}
Across the three benchmarks, performance trends vary across scaling and instruction tuning (\autoref{tab:main_result}, \autoref{fig:average_performance_across_category}). 
For AddOne and COMPCOMB, performance often decreases with instruction tuning and increased model size, whereas PLANE remains comparatively stable across variants. 
These findings indicate that improvements associated with scale and instruction tuning do not consistently correspond to improved compositional task performance under functional evaluation.

\subsection{Representational Evaluation}
\label{sec:compknow}

In our second evaluation setup, we examine compositionality at the level of internal representations. We conduct a layer-wise probing analysis to determine whether compositional signals are encoded across model architectures and how these patterns relate to functional task outcomes.

\paragraph{Methodology.}
For each model, we extract hidden states representing encoding of the task question, from every fifth transformer layer.  For AddOne and PLANE, we train a linear classifier on layer-wise representations to predict entailment labels. For COMPCOMB, we compute cosine similarity between token-level embeddings to assess type discrimination.  All results are compared against a random baseline (details in \autoref{appendix:c}).

\begin{figure}[!ht]
     \centering
     \begin{adjustbox}{minipage=\columnwidth,scale=1} 
     \begin{subfigure}[t]{0.49\columnwidth}
         \centering
         \includegraphics[width=\columnwidth]{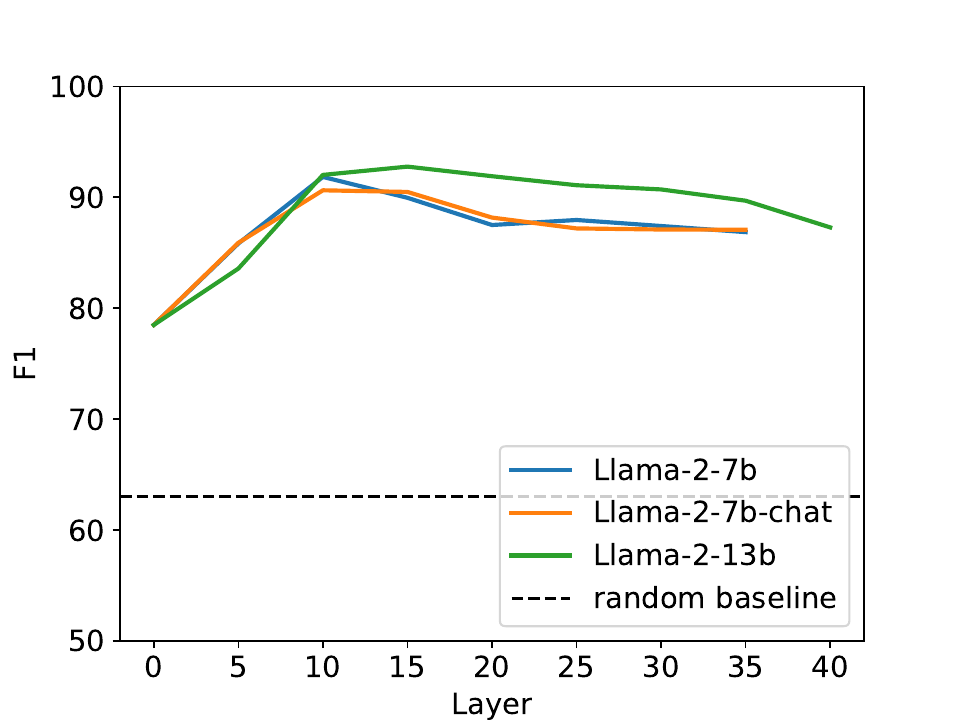}
         \caption{Llama2}
    \end{subfigure}
    \hfill
    \begin{subfigure}[t]{0.49\columnwidth}
         \centering
         \includegraphics[width=\columnwidth]{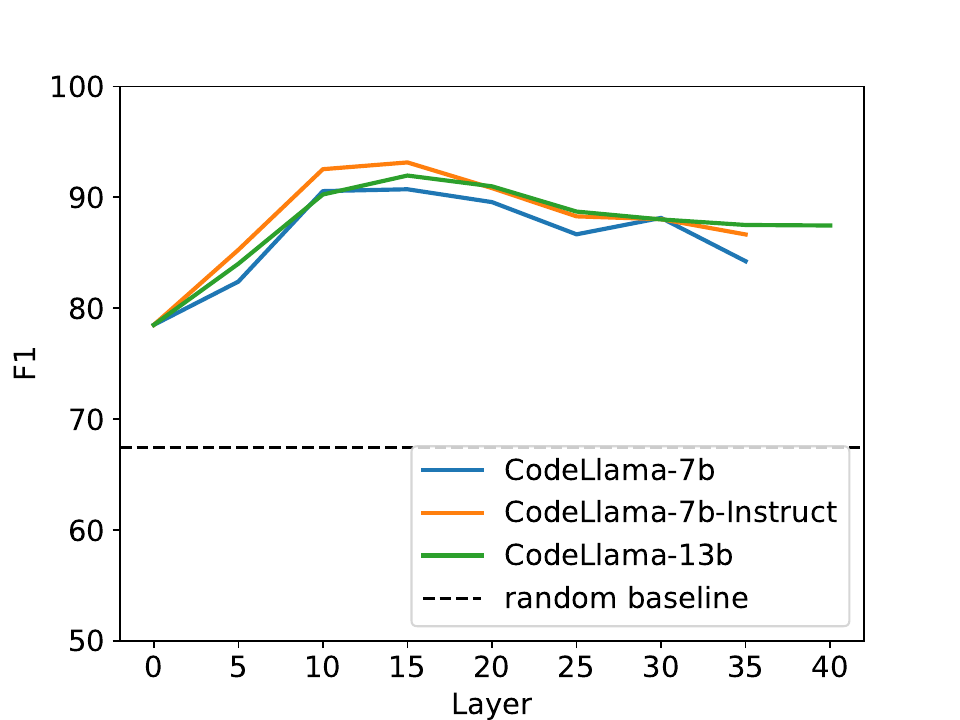}
         \caption{CodeLlama}
    \end{subfigure}
    \newline
     \begin{subfigure}[t]{0.49\columnwidth}
         \centering
         \includegraphics[width=\columnwidth]{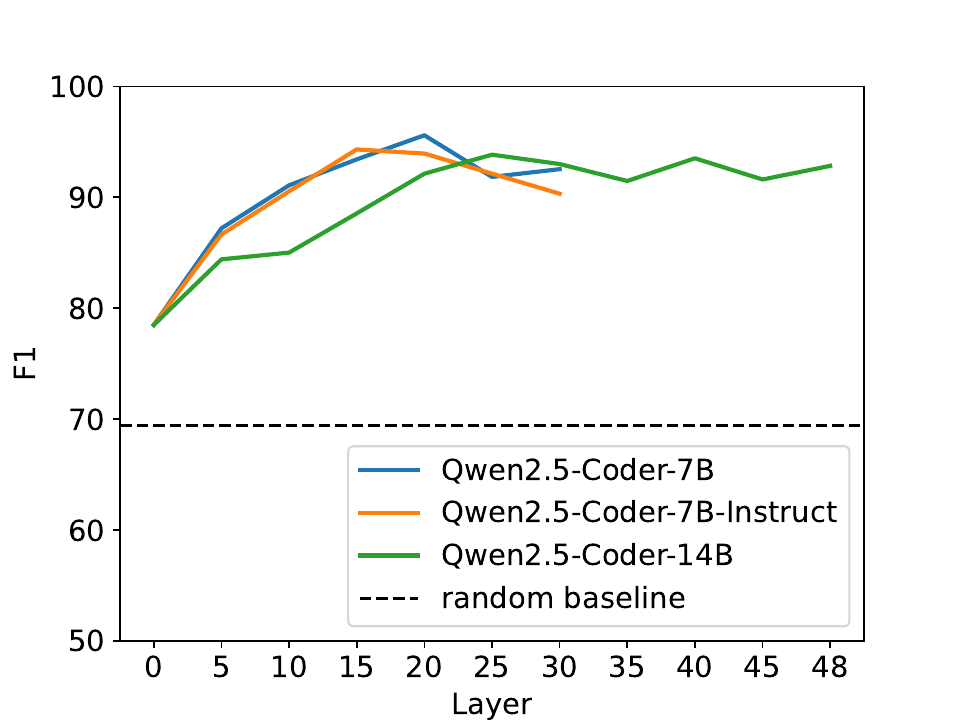}
         \caption{Qwen2.5-Coder}
    \end{subfigure}
    \hfill
    \begin{subfigure}[t]{0.49\columnwidth}
         \centering
         \includegraphics[width=\columnwidth]{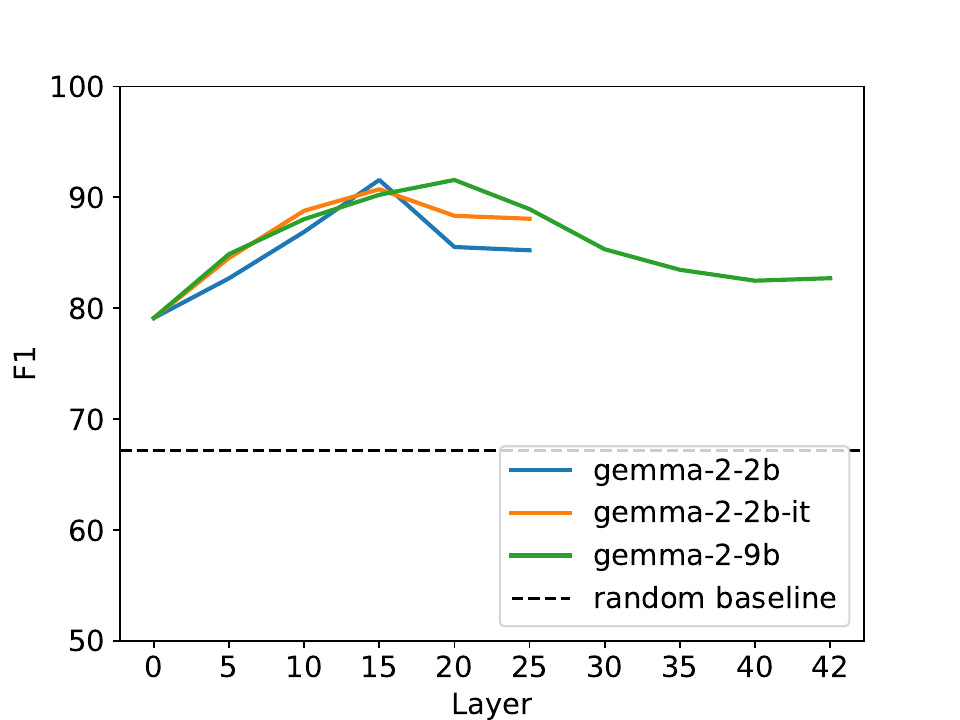}
         \caption{Gemma2}
    \end{subfigure}
\end{adjustbox}
\caption{Layer-wise results (weighted F1 score) on AddOne dataset.}
\label{fig:layer_wise_addone}
\end{figure}

\begin{figure}[!ht]
     \centering
     \begin{adjustbox}{minipage=\columnwidth,scale=1} 
     \begin{subfigure}[t]{0.49\columnwidth}
         \centering
         \includegraphics[width=\columnwidth]{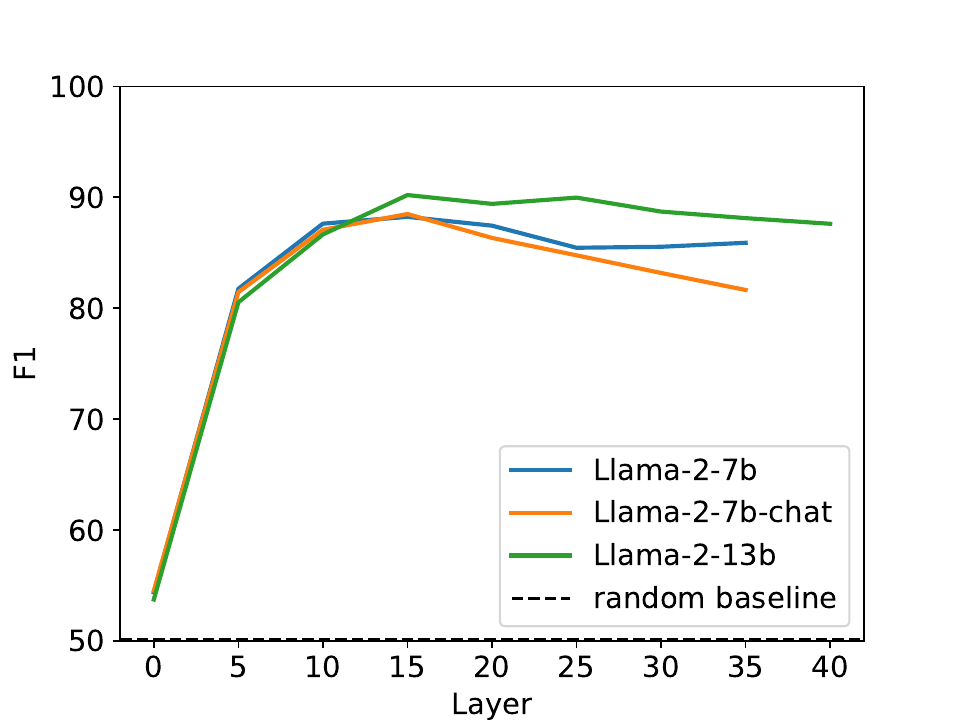}
         \caption{Llama2}
    \end{subfigure}
    \hfill
    \begin{subfigure}[t]{0.49\columnwidth}
         \centering
         \includegraphics[width=\columnwidth]{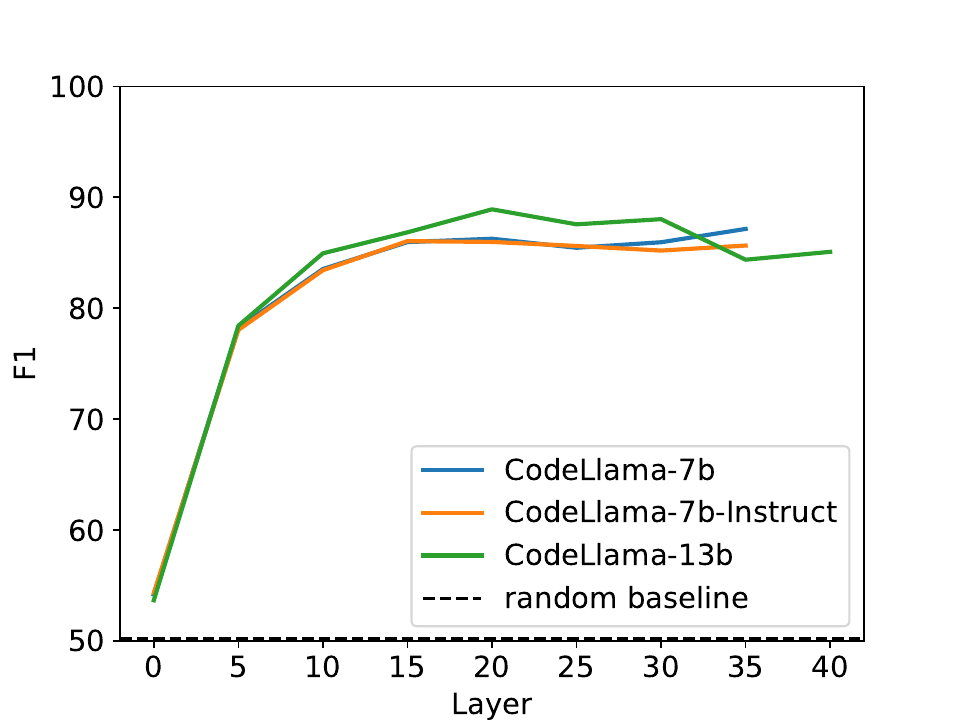}
         \caption{CodeLlama}
    \end{subfigure}
    \newline
     \begin{subfigure}[t]{0.49\columnwidth}
         \centering
         \includegraphics[width=\columnwidth]{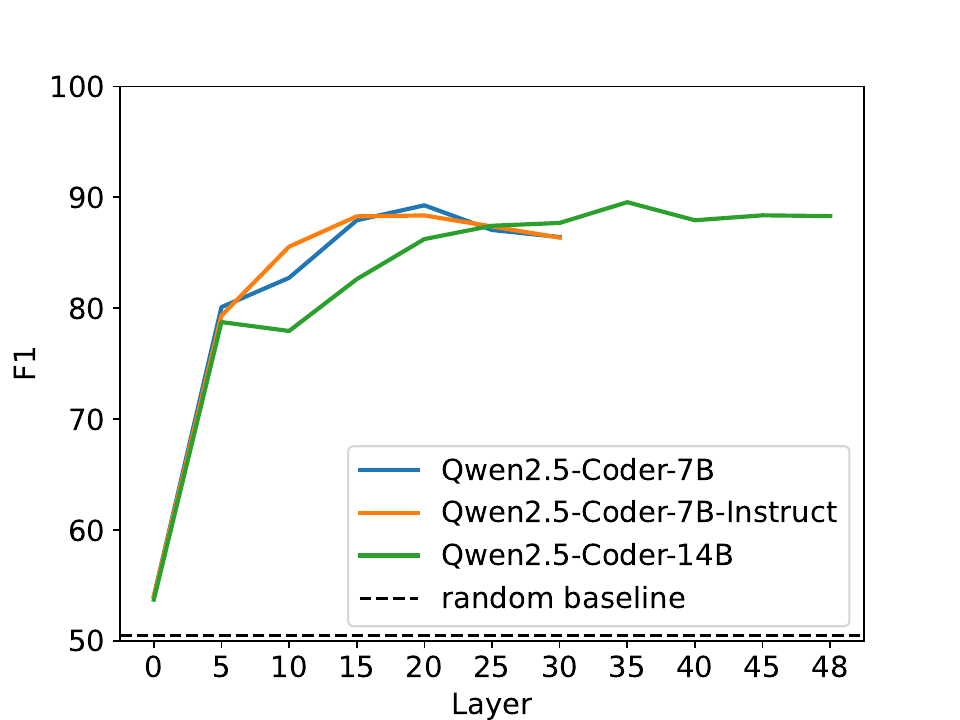}
         \caption{Qwen2.5-Coder}
    \end{subfigure}
    \hfill
    \begin{subfigure}[t]{0.49\columnwidth}
         \centering
         \includegraphics[width=\columnwidth]{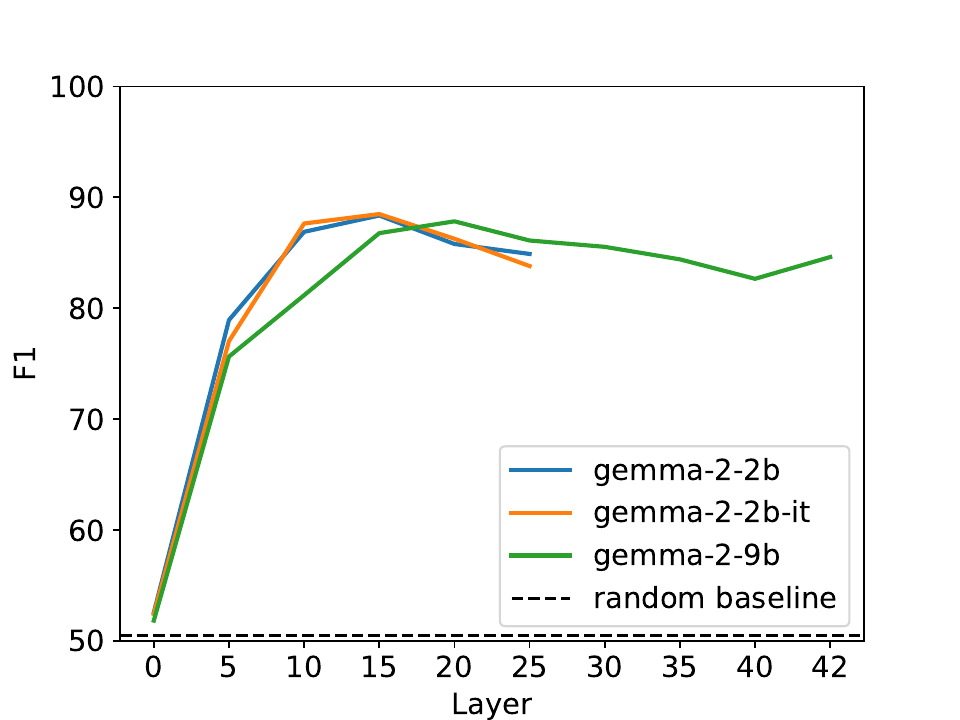}
         \caption{Gemma2}
    \end{subfigure}
\end{adjustbox}
\caption{Layer-wise results (weighted F1 score) on PLANE dataset.}
\label{fig:layer_wise_plane}
\end{figure}

\begin{figure}[!ht]
     \centering
     \begin{adjustbox}{minipage=\columnwidth,scale=1} 
     \begin{subfigure}[t]{0.49\columnwidth}
         \centering
         \includegraphics[width=\columnwidth]{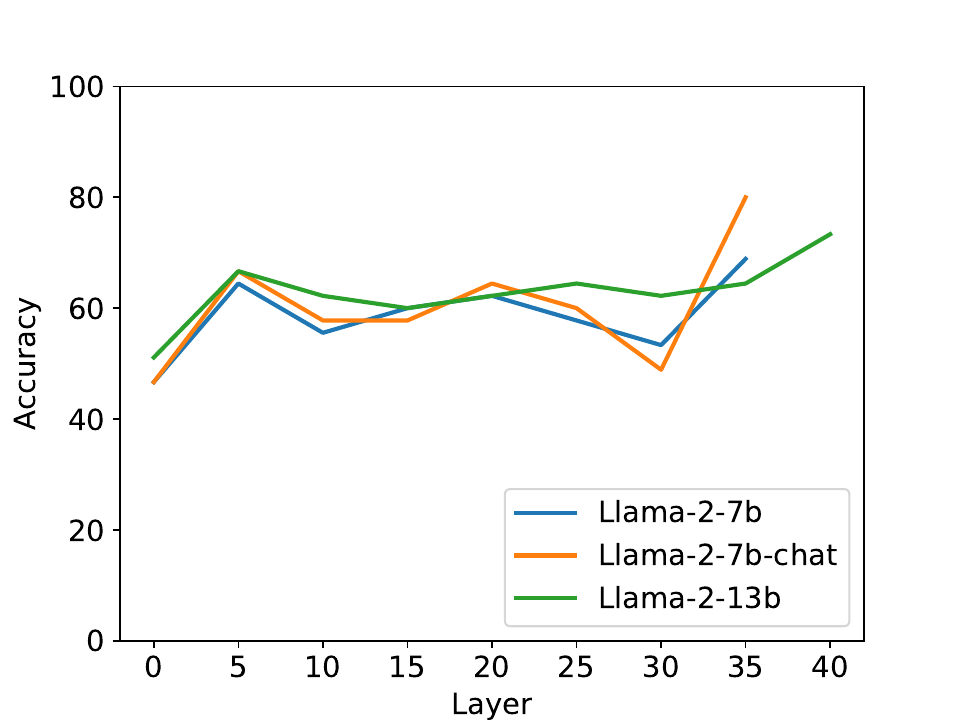}
         \caption{Llama2}
    \end{subfigure}
    \hfill
    \begin{subfigure}[t]{0.49\columnwidth}
         \centering
         \includegraphics[width=\columnwidth]{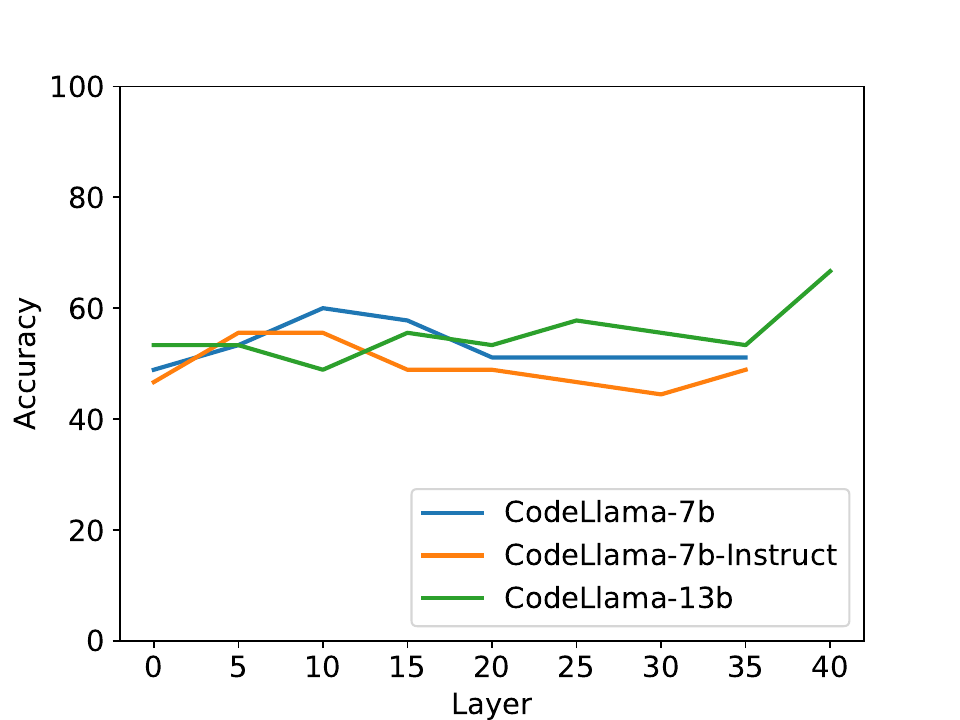}
         \caption{CodeLlama}
    \end{subfigure}
    \newline
     \begin{subfigure}[t]{0.49\columnwidth}
         \centering
         \includegraphics[width=\columnwidth]{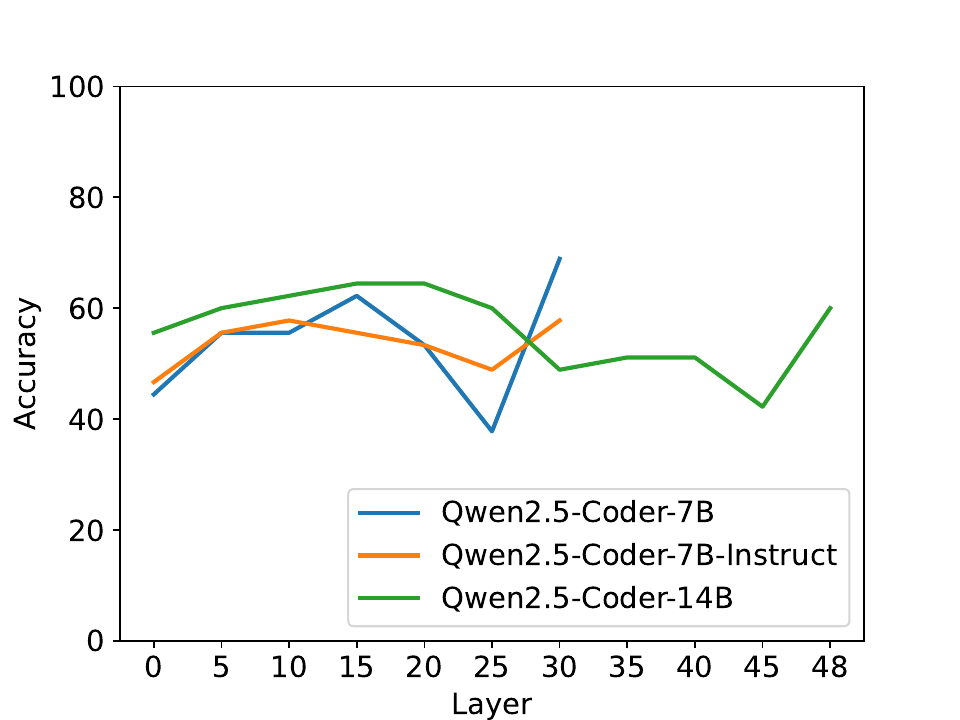}
         \caption{Qwen2.5-Coder}
    \end{subfigure}
    \hfill
    \begin{subfigure}[t]{0.49\columnwidth}
         \centering
         \includegraphics[width=\columnwidth]{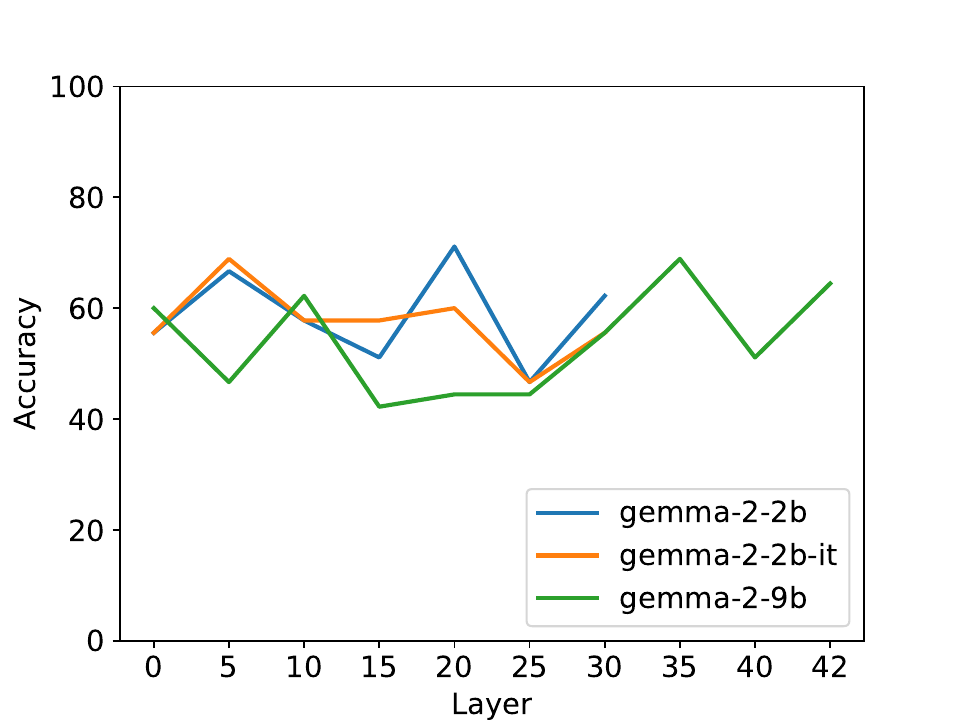}
         \caption{Gemma2}
    \end{subfigure}
\end{adjustbox}
\caption{Layer-wise results (accuracy) on CompComb dataset.}
\label{fig:layer_wise_compcomb}
\end{figure}

\paragraph{Results.}
Representations perform consistently above chance across models and layers (Figures~\ref{fig:layer_wise_addone}–\ref{fig:layer_wise_compcomb}), indicating that compositional information is encoded internally. 
For AddOne and PLANE, signal strength typically peaks in intermediate layers. 
Overall, representational evidence remains stable across scaling and instruction tuning.

\section{Conclusion}
\label{sec:disc}

In this work, we contrasted functional and representational evaluations of adjective–noun compositionality in LLMs. \textit{First}, we find that compositional task performance varies substantially across model variants. Although scaling and instruction tuning are typically associated with improved overall capability, they do not consistently improve compositional performance and in some cases lead to performance declines across benchmarks. This indicates that gains in general task performance do not uniformly translate to gains in compositional behavior under functional evaluation. \textit{Second}, representational evidence remains comparatively stable across model families and variants. Compositional structure is reliably encoded in internal representations, with signals often peaking in intermediate layers and exhibiting similar trends across base, instruction-tuned, and scaled models. Taken together, these findings highlight a systematic divergence between functional and representational perspectives on compositionality. Compositional knowledge appears to be consistently encoded in LLMs, yet its behavioral expression depends on the model variants and evaluation setup. Thus, evaluating models solely through task performance or internal analysis may yield incomplete conclusions; integrating functional and representational approaches offers a more comprehensive view. 


More broadly, these results underscore the importance of contrastive evaluation setups when assessing model capabilities. Relying solely on task-level performance or exclusively on representational probing may provide an incomplete picture of model behavior. For robustness and safety-sensitive deployment \cite{mahilraj2023evaluation, bender2021dangers}, understanding potential divergences between internal structure and external outputs is critical. A comprehensive evaluation strategy that integrates functional and representational analyses can therefore yield a more reliable and fine-grained account of model capabilities.

\section*{Limitations}
\label{sec:lim}

Our study has some limitations that constrain the scope of our conclusions. 
\textit{First}, our evaluation focuses exclusively on adjective–noun compositionality. While this domain provides a controlled and linguistically grounded testbed, compositionality manifests at multiple levels of linguistic structure. Extending this contrastive evaluation framework to subword, phrasal, and sentence-level composition, as well as to multilingual models, would enable broader generalization of our findings.

\textit{Second}, although we observe consistent representational signals, our analysis does not establish a causal relationship between internal compositional structure and task-level behavior. Layer-wise probing reveals the presence of information, but does not determine whether or how it is functionally used during inference. Future work could employ causal intervention methods—such as activation patching or path attribution—to test whether compositional representations directly influence model predictions.

\textit{Finally}, our comparison is limited to two evaluation paradigms: prompt-based functional assessment and linear representational probing. Developing more comprehensive contrastive evaluation frameworks will be important for understanding how different analytical lenses shape conclusions about compositionality and other model capabilities.

\section*{Ethical Considerations}
We do not anticipate any risks in our work. Any data used does not contain any personal details or offensive content. In this study, our use of all artifacts is consistent with the license and intended use of all datasets and models. The AddOne Dataset \cite{pavlick-callison-burch-2016-babies} is available on HuggingFace on\href{https://huggingface.co/datasets/pietrolesci/add_one_rte/edit/main/README.md}{this page} and is free for all use. The PLANE dataset \cite{bertolini2022testing} is also available on HuggingFace on\href{https://huggingface.co/datasets/lorenzoscottb/PLANE-ood}{this page} and has license of Creative Commons Attribution 2.0. Our toy dataset COMPCOMB will be released upon publication along with details of license and intended use.  All models used are also from HuggingFace and their details, along with licenses, are summarized  in \autoref{appendix:b}. All code and data will be released upon publication for greater reproducibility.

\bibliography{custom}

\begin{thebibliography}{64}
\providecommand{\natexlab}[1]{#1}

\bibitem[{Bender et~al.(2021)Bender, Gebru, McMillan-Major, and Shmitchell}]{bender2021dangers}
Emily~M Bender, Timnit Gebru, Angelina McMillan-Major, and Shmargaret Shmitchell. 2021.
\newblock On the dangers of stochastic parrots: Can language models be too big?
\newblock In \emph{Proceedings of the 2021 ACM conference on fairness, accountability, and transparency}, pages 610--623.

\bibitem[{Bertolini et~al.(2022)Bertolini, Weeds, and Weir}]{bertolini2022testing}
Lorenzo Bertolini, Julie Weeds, and David Weir. 2022.
\newblock Testing large language models on compositionality and inference with phrase-level adjective-noun entailment.
\newblock In \emph{Proceedings of the 29th International Conference on Computational Linguistics}, pages 4084--4100.

\bibitem[{Bertolini et~al.(2021)Bertolini, Weeds, Weir, and Peng}]{bertolini-etal-2021-representing}
Lorenzo Bertolini, Julie Weeds, David Weir, and Qiwei Peng. 2021.
\newblock \href {https://doi.org/10.18653/v1/2021.findings-acl.296} {Representing syntax and composition with geometric transformations}.
\newblock In \emph{Findings of the Association for Computational Linguistics: ACL-IJCNLP 2021}, pages 3343--3353, Online. Association for Computational Linguistics.

\bibitem[{Biderman et~al.(2023)Biderman, Schoelkopf, Anthony, Bradley, O’Brien, Hallahan, Khan, Purohit, Prashanth, Raff et~al.}]{biderman2023pythia}
Stella Biderman, Hailey Schoelkopf, Quentin~Gregory Anthony, Herbie Bradley, Kyle O’Brien, Eric Hallahan, Mohammad~Aflah Khan, Shivanshu Purohit, USVSN~Sai Prashanth, Edward Raff, and 1 others. 2023.
\newblock Pythia: A suite for analyzing large language models across training and scaling.
\newblock In \emph{International Conference on Machine Learning}, pages 2397--2430. PMLR.

\bibitem[{Block(1981)}]{Block1981-BLOPAB}
Ned Block. 1981.
\newblock \href {https://doi.org/10.2307/2184371} {Psychologism and behaviorism}.
\newblock \emph{Philosophical Review}, 90(1):5--43.

\bibitem[{Carnap(1988)}]{carnap1988meaning}
Rudolf Carnap. 1988.
\newblock \emph{Meaning and necessity: A study in semantics and modal logic}, volume~30.
\newblock University of Chicago Press.

\bibitem[{Chen et~al.(2024)Chen, Pan, Yu, Song, Wang, Yu, and Chen}]{chen2024skillsincontext}
Jiaao Chen, Xiaoman Pan, Dian Yu, Kaiqiang Song, Xiaoyang Wang, Dong Yu, and Jianshu Chen. 2024.
\newblock \href {https://openreview.net/forum?id=s1ByDEbpI8} {Skills-in-context prompting: Unlocking compositionality in large language models}.

\bibitem[{Chomsky(1956)}]{chomsky1956three}
Noam Chomsky. 1956.
\newblock Three models for the description of language.
\newblock \emph{IRE Transactions on information theory}, 2(3):113--124.

\bibitem[{Chowdhery et~al.(2023)Chowdhery, Narang, Devlin, Bosma, Mishra, Roberts, Barham, Chung, Sutton, Gehrmann et~al.}]{chowdhery2023palm}
Aakanksha Chowdhery, Sharan Narang, Jacob Devlin, Maarten Bosma, Gaurav Mishra, Adam Roberts, Paul Barham, Hyung~Won Chung, Charles Sutton, Sebastian Gehrmann, and 1 others. 2023.
\newblock Palm: Scaling language modeling with pathways.
\newblock \emph{Journal of Machine Learning Research}, 24(240):1--113.

\bibitem[{Chung et~al.(2024)Chung, Hou, Longpre, Zoph, Tay, Fedus, Li, Wang, Dehghani, Brahma et~al.}]{chung2024scaling}
Hyung~Won Chung, Le~Hou, Shayne Longpre, Barret Zoph, Yi~Tay, William Fedus, Yunxuan Li, Xuezhi Wang, Mostafa Dehghani, Siddhartha Brahma, and 1 others. 2024.
\newblock Scaling instruction-finetuned language models.
\newblock \emph{Journal of Machine Learning Research}, 25(70):1--53.

\bibitem[{Csord{\'a}s(2023)}]{csordas2023systematic}
R{\'o}bert Csord{\'a}s. 2023.
\newblock Systematic generalization in connectionist models.

\bibitem[{Cummins(1996)}]{cummins1996representations}
Robert Cummins. 1996.
\newblock \emph{Representations, targets, and attitudes}.
\newblock MIT press.

\bibitem[{Donatelli and Koller(2023)}]{donatelli2023compositionality}
Lucia Donatelli and Alexander Koller. 2023.
\newblock Compositionality in computational linguistics.
\newblock \emph{Annual Review of Linguistics}, 9(1):463--481.

\bibitem[{Drozdov et~al.(2023)Drozdov, Sch{\"a}rli, Aky{\"u}rek, Scales, Song, Chen, Bousquet, and Zhou}]{drozdov2023compositional}
Andrew Drozdov, Nathanael Sch{\"a}rli, Ekin Aky{\"u}rek, Nathan Scales, Xinying Song, Xinyun Chen, Olivier Bousquet, and Denny Zhou. 2023.
\newblock \href {https://openreview.net/forum?id=gJW8hSGBys8} {Compositional semantic parsing with large language models}.
\newblock In \emph{The Eleventh International Conference on Learning Representations}.

\bibitem[{Dziri et~al.(2024)Dziri, Lu, Sclar, Li, Jiang, Lin, Welleck, West, Bhagavatula, Le~Bras et~al.}]{dziri2024faith}
Nouha Dziri, Ximing Lu, Melanie Sclar, Xiang~Lorraine Li, Liwei Jiang, Bill~Yuchen Lin, Sean Welleck, Peter West, Chandra Bhagavatula, Ronan Le~Bras, and 1 others. 2024.
\newblock Faith and fate: Limits of transformers on compositionality.
\newblock \emph{Advances in Neural Information Processing Systems}, 36.

\bibitem[{Fodor and Pylyshyn(1988)}]{fodor1988connectionism}
Jerry~A Fodor and Zenon~W Pylyshyn. 1988.
\newblock Connectionism and cognitive architecture: A critical analysis.
\newblock \emph{Cognition}, 28(1-2):3--71.

\bibitem[{Geiger et~al.(2019)Geiger, Cases, Karttunen, and Potts}]{geiger-etal-2019-posing}
Atticus Geiger, Ignacio Cases, Lauri Karttunen, and Christopher Potts. 2019.
\newblock \href {https://doi.org/10.18653/v1/D19-1456} {Posing fair generalization tasks for natural language inference}.
\newblock In \emph{Proceedings of the 2019 Conference on Empirical Methods in Natural Language Processing and the 9th International Joint Conference on Natural Language Processing (EMNLP-IJCNLP)}, pages 4485--4495, Hong Kong, China. Association for Computational Linguistics.

\bibitem[{Guevara(2010)}]{guevara2010regression}
Emiliano~Raul Guevara. 2010.
\newblock A regression model of adjective-noun compositionality in distributional semantics.
\newblock In \emph{Proceedings of the 2010 workshop on geometrical models of natural language semantics}, pages 33--37.

\bibitem[{Guo et~al.(2025)Guo, Xue, Xu, Bo, Ye, Pierrehumbert, and Lewis}]{guo2025quantifying}
Zhijin Guo, Chenhao Xue, Zhaozhen Xu, Hongbo Bo, Yuxuan Ye, Janet~B Pierrehumbert, and Martha Lewis. 2025.
\newblock Quantifying compositionality of classic and state-of-the-art embeddings.
\newblock In \emph{Findings of the Association for Computational Linguistics: EMNLP 2025}, pages 22130--22146.

\bibitem[{Hartung et~al.(2017)Hartung, Kaupmann, Jebbara, and Cimiano}]{hartung2017learning}
Matthias Hartung, Fabian Kaupmann, Soufian Jebbara, and Philipp Cimiano. 2017.
\newblock Learning compositionality functions on word embeddings for modelling attribute meaning in adjective-noun phrases.
\newblock In \emph{Proceedings of the 15th Conference of the European Chapter of the Association for Computational Linguistics: Volume 1, Long Papers}, pages 54--64.

\bibitem[{Hewitt and Manning(2019)}]{hewitt-manning-2019-structural}
John Hewitt and Christopher~D. Manning. 2019.
\newblock \href {https://doi.org/10.18653/v1/N19-1419} {{A} structural probe for finding syntax in word representations}.
\newblock In \emph{Proceedings of the 2019 Conference of the North {A}merican Chapter of the Association for Computational Linguistics: Human Language Technologies, Volume 1 (Long and Short Papers)}, pages 4129--4138, Minneapolis, Minnesota. Association for Computational Linguistics.

\bibitem[{Hoffmann et~al.(2022)Hoffmann, Borgeaud, Mensch, Buchatskaya, Cai, Rutherford, Casas, Hendricks, Welbl, Clark et~al.}]{hoffmann2022training}
Jordan Hoffmann, Sebastian Borgeaud, Arthur Mensch, Elena Buchatskaya, Trevor Cai, Eliza Rutherford, Diego de~Las Casas, Lisa~Anne Hendricks, Johannes Welbl, Aidan Clark, and 1 others. 2022.
\newblock Training compute-optimal large language models.
\newblock \emph{arXiv preprint arXiv:2203.15556}.

\bibitem[{Hupkes et~al.(2020)Hupkes, Dankers, Mul, and Bruni}]{hupkes2020compositionality}
Dieuwke Hupkes, Verna Dankers, Mathijs Mul, and Elia Bruni. 2020.
\newblock Compositionality decomposed: How do neural networks generalise?
\newblock \emph{Journal of Artificial Intelligence Research}, 67:757--795.

\bibitem[{Hupkes et~al.(2023)Hupkes, Giulianelli, Dankers, Artetxe, Elazar, Pimentel, Christodoulopoulos, Lasri, Saphra, Sinclair et~al.}]{hupkes2023taxonomy}
Dieuwke Hupkes, Mario Giulianelli, Verna Dankers, Mikel Artetxe, Yanai Elazar, Tiago Pimentel, Christos Christodoulopoulos, Karim Lasri, Naomi Saphra, Arabella Sinclair, and 1 others. 2023.
\newblock A taxonomy and review of generalization research in nlp.
\newblock \emph{Nature Machine Intelligence}, 5(10):1161--1174.

\bibitem[{Kamp and Partee(1995)}]{kamp1995prototype}
Hans Kamp and Barbara Partee. 1995.
\newblock Prototype theory and compositionality.
\newblock \emph{Cognition}, 57(2):129--191.

\bibitem[{Kaplan et~al.(2020)Kaplan, McCandlish, Henighan, Brown, Chess, Child, Gray, Radford, Wu, and Amodei}]{kaplan2020scaling}
Jared Kaplan, Sam McCandlish, Tom Henighan, Tom~B Brown, Benjamin Chess, Rewon Child, Scott Gray, Alec Radford, Jeffrey Wu, and Dario Amodei. 2020.
\newblock Scaling laws for neural language models.
\newblock \emph{arXiv preprint arXiv:2001.08361}.

\bibitem[{Kim and Linzen(2020)}]{kim-linzen-2020-cogs}
Najoung Kim and Tal Linzen. 2020.
\newblock \href {https://doi.org/10.18653/v1/2020.emnlp-main.731} {{COGS}: A compositional generalization challenge based on semantic interpretation}.
\newblock In \emph{Proceedings of the 2020 Conference on Empirical Methods in Natural Language Processing (EMNLP)}, pages 9087--9105, Online. Association for Computational Linguistics.

\bibitem[{Kumon and Yanaka(2025)}]{kumon-yanaka-2025-analyzing}
Ryoma Kumon and Hitomi Yanaka. 2025.
\newblock \href {https://doi.org/10.18653/v1/2025.naacl-long.432} {Analyzing the inner workings of transformers in compositional generalization}.
\newblock In \emph{Proceedings of the 2025 Conference of the Nations of the Americas Chapter of the Association for Computational Linguistics: Human Language Technologies (Volume 1: Long Papers)}, pages 8529--8540, Albuquerque, New Mexico. Association for Computational Linguistics.

\bibitem[{Lake and Baroni(2018)}]{lake2018generalization}
Brenden Lake and Marco Baroni. 2018.
\newblock Generalization without systematicity: On the compositional skills of sequence-to-sequence recurrent networks.
\newblock In \emph{International conference on machine learning}, pages 2873--2882. PMLR.

\bibitem[{Lampinen et~al.(2024)Lampinen, Chan, and Hermann}]{lampinen2024learned}
Andrew~Kyle Lampinen, Stephanie~C.Y. Chan, and Katherine Hermann. 2024.
\newblock \href {https://openreview.net/forum?id=aY2nsgE97a} {Learned feature representations are biased by complexity, learning order, position, and more}.
\newblock \emph{Transactions on Machine Learning Research}.

\bibitem[{Lebowitz(1983)}]{lebowitz1983generalization}
Michael Lebowitz. 1983.
\newblock Generalization from natural language text.
\newblock \emph{Cognitive Science}, 7(1):1--40.

\bibitem[{Li et~al.(2024{\natexlab{a}})Li, Li, Jing, Wu, Zhai, and Jia}]{li2024compositional}
Chuanhao Li, Zhen Li, Chenchen Jing, Yuwei Wu, Mingliang Zhai, and Yunde Jia. 2024{\natexlab{a}}.
\newblock Compositional substitutivity of visual reasoning for visual question answering.
\newblock In \emph{European Conference on Computer Vision}, pages 143--160. Springer.

\bibitem[{Li et~al.(2024{\natexlab{b}})Li, Jiang, Xie, Song, Lian, and Wei}]{li2024understanding}
Zhaoyi Li, Gangwei Jiang, Hong Xie, Linqi Song, Defu Lian, and Ying Wei. 2024{\natexlab{b}}.
\newblock Understanding and patching compositional reasoning in llms.
\newblock \emph{arXiv preprint arXiv:2402.14328}.

\bibitem[{Longpre et~al.(2023)Longpre, Hou, Vu, Webson, Chung, Tay, Zhou, Le, Zoph, Wei et~al.}]{longpre2023flan}
Shayne Longpre, Le~Hou, Tu~Vu, Albert Webson, Hyung~Won Chung, Yi~Tay, Denny Zhou, Quoc~V Le, Barret Zoph, Jason Wei, and 1 others. 2023.
\newblock The flan collection: Designing data and methods for effective instruction tuning.
\newblock In \emph{International Conference on Machine Learning}, pages 22631--22648. PMLR.

\bibitem[{Lu et~al.(2024{\natexlab{a}})Lu, Peng, Cheng, Galley, Chang, Wu, Zhu, and Gao}]{lu2024chameleon}
Pan Lu, Baolin Peng, Hao Cheng, Michel Galley, Kai-Wei Chang, Ying~Nian Wu, Song-Chun Zhu, and Jianfeng Gao. 2024{\natexlab{a}}.
\newblock Chameleon: Plug-and-play compositional reasoning with large language models.
\newblock \emph{Advances in Neural Information Processing Systems}, 36.

\bibitem[{Lu et~al.(2024{\natexlab{b}})Lu, Schuff, and Gurevych}]{lu2024prompts}
Sheng Lu, Hendrik Schuff, and Iryna Gurevych. 2024{\natexlab{b}}.
\newblock How are prompts different in terms of sensitivity?
\newblock In \emph{Proceedings of the 2024 Conference of the North American Chapter of the Association for Computational Linguistics: Human Language Technologies (Volume 1: Long Papers)}, pages 5833--5856.

\bibitem[{Ma et~al.(2024)Ma, Li, and Liang}]{ma2024examination}
Teli Ma, Rong Li, and Junwei Liang. 2024.
\newblock An examination of the compositionality of large generative vision-language models.
\newblock In \emph{Proceedings of the 2024 Conference of the North American Chapter of the Association for Computational Linguistics: Human Language Technologies (Volume 1: Long Papers)}, pages 692--705.

\bibitem[{Macken(2014)}]{macken2014representation}
Marlys~A Macken. 2014.
\newblock Representation, rules, and overgeneralization in phonology.
\newblock In \emph{Mechanisms of language acquisition}, pages 367--397. Psychology Press.

\bibitem[{Mahilraj et~al.(2023)Mahilraj, Pandian, Subbiah, Kalyan et~al.}]{mahilraj2023evaluation}
Jenifer Mahilraj, Manivel Pandian, Muthuraman Subbiah, S~Kalyan, and 1 others. 2023.
\newblock Evaluation of the robustness, transparency, reliability and safety of ai systems.
\newblock In \emph{2023 9th International Conference on Advanced Computing and Communication Systems (ICACCS)}, volume~1, pages 2526--2535. IEEE.

\bibitem[{Mahowald et~al.(2024)Mahowald, Ivanova, Blank, Kanwisher, Tenenbaum, and Fedorenko}]{mahowald2024dissociating}
Kyle Mahowald, Anna~A Ivanova, Idan~A Blank, Nancy Kanwisher, Joshua~B Tenenbaum, and Evelina Fedorenko. 2024.
\newblock Dissociating language and thought in large language models.
\newblock \emph{Trends in Cognitive Sciences}.

\bibitem[{McCurdy et~al.(2024)McCurdy, Soulos, Smolensky, Fernandez, and Gao}]{mccurdy-etal-2024-toward}
Kate McCurdy, Paul Soulos, Paul Smolensky, Roland Fernandez, and Jianfeng Gao. 2024.
\newblock \href {https://doi.org/10.18653/v1/2024.emnlp-main.524} {Toward compositional behavior in neural models: A survey of current views}.
\newblock In \emph{Proceedings of the 2024 Conference on Empirical Methods in Natural Language Processing}, pages 9323--9339, Miami, Florida, USA. Association for Computational Linguistics.

\bibitem[{Min et~al.(2023)Min, Ross, Sulem, Veyseh, Nguyen, Sainz, Agirre, Heintz, and Roth}]{10.1145/3605943}
Bonan Min, Hayley Ross, Elior Sulem, Amir Pouran~Ben Veyseh, Thien~Huu Nguyen, Oscar Sainz, Eneko Agirre, Ilana Heintz, and Dan Roth. 2023.
\newblock \href {https://doi.org/10.1145/3605943} {Recent advances in natural language processing via large pre-trained language models: A survey}.
\newblock \emph{ACM Comput. Surv.}, 56(2).

\bibitem[{Murty et~al.(2023)Murty, Sharma, Andreas, and Manning}]{murty2023characterizing}
Shikhar Murty, Pratyusha Sharma, Jacob Andreas, and Christopher~D Manning. 2023.
\newblock \href {https://openreview.net/forum?id=sAOOeI878Ns} {Characterizing intrinsic compositionality in transformers with tree projections}.
\newblock In \emph{The Eleventh International Conference on Learning Representations}.

\bibitem[{Orgad et~al.(2025)Orgad, Toker, Gekhman, Reichart, Szpektor, Kotek, and Belinkov}]{orgad2025llms}
Hadas Orgad, Michael Toker, Zorik Gekhman, Roi Reichart, Idan Szpektor, Hadas Kotek, and Yonatan Belinkov. 2025.
\newblock \href {https://openreview.net/forum?id=KRnsX5Em3W} {{LLM}s know more than they show: On the intrinsic representation of {LLM} hallucinations}.
\newblock In \emph{The Thirteenth International Conference on Learning Representations}.

\bibitem[{Ouyang et~al.(2022)Ouyang, Wu, Jiang, Almeida, Wainwright, Mishkin, Zhang, Agarwal, Slama, Ray et~al.}]{ouyang2022training}
Long Ouyang, Jeffrey Wu, Xu~Jiang, Diogo Almeida, Carroll Wainwright, Pamela Mishkin, Chong Zhang, Sandhini Agarwal, Katarina Slama, Alex Ray, and 1 others. 2022.
\newblock Training language models to follow instructions with human feedback.
\newblock \emph{Advances in neural information processing systems}, 35:27730--27744.

\bibitem[{Pagin and Westerst{\aa}hl(2010{\natexlab{a}})}]{pagin2010compositionality}
Peter Pagin and Dag Westerst{\aa}hl. 2010{\natexlab{a}}.
\newblock Compositionality ii: Arguments and problems.
\newblock \emph{Philosophy Compass}, 5(3):265--282.

\bibitem[{Pagin and Westerst{\aa}hl(2010{\natexlab{b}})}]{pagin2010pure}
Peter Pagin and Dag Westerst{\aa}hl. 2010{\natexlab{b}}.
\newblock Pure quotation and general compositionality.
\newblock \emph{Linguistics and Philosophy}, 33:381--415.

\bibitem[{Partee et~al.(1995)}]{partee1995lexical}
Barbara Partee and 1 others. 1995.
\newblock Lexical semantics and compositionality.
\newblock \emph{An invitation to cognitive science: Language}, 1:311--360.

\bibitem[{Pavlick and Callison-Burch(2016)}]{pavlick-callison-burch-2016-babies}
Ellie Pavlick and Chris Callison-Burch. 2016.
\newblock \href {https://doi.org/10.18653/v1/P16-1204} {Most {``}babies{''} are {``}little{''} and most {``}problems{''} are {``}huge{''}: Compositional entailment in adjective-nouns}.
\newblock In \emph{Proceedings of the 54th Annual Meeting of the Association for Computational Linguistics (Volume 1: Long Papers)}, pages 2164--2173, Berlin, Germany. Association for Computational Linguistics.

\bibitem[{Peng et~al.(2025)Peng, Chai, and S{\o}gaard}]{peng2025understanding}
Qiwei Peng, Yekun Chai, and Anders S{\o}gaard. 2025.
\newblock Understanding subword compositionality of large language models.
\newblock In \emph{Proceedings of the 2025 Conference on Empirical Methods in Natural Language Processing}, pages 22535--22546.

\bibitem[{Pickel and Szab{\'o}(2004)}]{pickel2004compositionality}
Bryan Pickel and Zolt{\'a}n~Gendler Szab{\'o}. 2004.
\newblock Compositionality.

\bibitem[{Press et~al.(2023)Press, Zhang, Min, Schmidt, Smith, and Lewis}]{press-etal-2023-measuring}
Ofir Press, Muru Zhang, Sewon Min, Ludwig Schmidt, Noah Smith, and Mike Lewis. 2023.
\newblock \href {https://doi.org/10.18653/v1/2023.findings-emnlp.378} {Measuring and narrowing the compositionality gap in language models}.
\newblock In \emph{Findings of the Association for Computational Linguistics: EMNLP 2023}, pages 5687--5711, Singapore. Association for Computational Linguistics.

\bibitem[{Pullum and Scholz(2010)}]{pullum2010recursion}
Geoffrey~K Pullum and Barbara~C Scholz. 2010.
\newblock Recursion and the infinitude claim.
\newblock In \emph{Recursion and human language}, pages 113--137. De Gruyter Mouton.

\bibitem[{Saba(2023)}]{saba2023stochastic}
Walid~S Saba. 2023.
\newblock Stochastic llms do not understand language: towards symbolic, explainable and ontologically based llms.
\newblock In \emph{International conference on conceptual modeling}, pages 3--19. Springer.

\bibitem[{Sclar et~al.(2024)Sclar, Choi, Tsvetkov, and Suhr}]{sclar2024quantifying}
Melanie Sclar, Yejin Choi, Yulia Tsvetkov, and Alane Suhr. 2024.
\newblock \href {https://openreview.net/forum?id=RIu5lyNXjT} {Quantifying language models' sensitivity to spurious features in prompt design or: How i learned to start worrying about prompt formatting}.
\newblock In \emph{The Twelfth International Conference on Learning Representations}.

\bibitem[{SHAO et~al.(2023)SHAO, Cai, xu, Liao, Zheng, and Yang}]{shao2023compositional}
NAN SHAO, Zefan Cai, Hanwei xu, Chonghua Liao, Yanan Zheng, and Zhilin Yang. 2023.
\newblock \href {https://openreview.net/forum?id=6axIMJA7ME3} {Compositional task representations for large language models}.
\newblock In \emph{The Eleventh International Conference on Learning Representations}.

\bibitem[{Smolensky(1987)}]{smolensky1987connectionist}
Paul Smolensky. 1987.
\newblock Connectionist ai, symbolic ai, and the brain.
\newblock \emph{Artificial Intelligence Review}, 1(2):95--109.

\bibitem[{Sun et~al.(2024)Sun, Shaib, and Wallace}]{sun2024evaluating}
Jiuding Sun, Chantal Shaib, and Byron~C Wallace. 2024.
\newblock \href {https://openreview.net/forum?id=g9diuvxN6D} {Evaluating the zero-shot robustness of instruction-tuned language models}.
\newblock In \emph{The Twelfth International Conference on Learning Representations}.

\bibitem[{Symons and Calvo(2014)}]{symons2014systematicity}
John Symons and Paco Calvo. 2014.
\newblock Systematicity: an overview.
\newblock \emph{P. Calvo \& J. Symons,(Eds.)}.

\bibitem[{Townsend et~al.(2018)Townsend, Engesser, Stoll, Zuberb{\"u}hler, and Bickel}]{townsend2018compositionality}
Simon~W Townsend, Sabrina Engesser, Sabine Stoll, Klaus Zuberb{\"u}hler, and Balthasar Bickel. 2018.
\newblock Compositionality in animals and humans.
\newblock \emph{PLoS Biology}, 16(8):e2006425.

\bibitem[{Wei et~al.(2022)Wei, Bosma, Zhao, Guu, Yu, Lester, Du, Dai, and Le}]{wei2022finetuned}
Jason Wei, Maarten Bosma, Vincent Zhao, Kelvin Guu, Adams~Wei Yu, Brian Lester, Nan Du, Andrew~M. Dai, and Quoc~V Le. 2022.
\newblock \href {https://openreview.net/forum?id=gEZrGCozdqR} {Finetuned language models are zero-shot learners}.
\newblock In \emph{International Conference on Learning Representations}.

\bibitem[{Werning et~al.(2012)Werning, Hinzen, and Machery}]{werning2012oxford}
Markus Werning, Wolfram Hinzen, and Edouard Machery. 2012.
\newblock \emph{The Oxford handbook of compositionality}.
\newblock Oxford University Press.

\bibitem[{Xu et~al.(2024)Xu, Shi, and Liang}]{xu2024large}
Zhuoyan Xu, Zhenmei Shi, and Yingyu Liang. 2024.
\newblock Do large language models have compositional ability? an investigation into limitations and scalability.
\newblock In \emph{ICLR 2024 Workshop on Mathematical and Empirical Understanding of Foundation Models}.

\bibitem[{Zhang et~al.(2024)Zhang, He, Lei, Yue, Wang, and Lu}]{zhang2024can}
Min Zhang, Jianfeng He, Shuo Lei, Murong Yue, Linhan Wang, and Chang-Tien Lu. 2024.
\newblock Can llm find the green circle? investigation and human-guided tool manipulation for compositional generalization.
\newblock In \emph{ICASSP 2024-2024 IEEE International Conference on Acoustics, Speech and Signal Processing (ICASSP)}, pages 11996--12000. IEEE.

\end{thebibliography}

\appendix

\section{Tasks and Datasets}
\label{sec:appendix}

In this appendix section, we provide a detailed explanation of the different aspects of compositionality we evaluate \cite{hupkes2020compositionality} along with the task used.

\textbf{Substitutivity: } It refers to a model’s ability to preserve meaning when a component of a complex expression is replaced with a semantically equivalent or synonymous alternative \cite{carnap1988meaning, hupkes2020compositionality}. In a truly compositional system, meaning should be a function of the meanings of its parts and their combination -- so substituting one part with a synonym should not alter the overall meaning of the expression. We evaluate this using the \textbf{AddOne Dataset} \cite{pavlick-callison-burch-2016-babies}, which tests whether models recognize entailment relationships between original and modified sentences. For example, given the sentence \textit{The runner set a record}, we substitute the noun \textit{record} with an adjective–noun phrase such as \textit{new record}, producing the modified sentence \textit{The runner set a new record}. While the surface form changes, the core meaning remains unchanged in context. A model with compositional understanding should therefore recognize the entailment:
\begin{center}
\textit{The runner set a new record} $\models$ \textit{The runner set a record}.
\end{center}

Success on this task requires more than detecting synonymy at the word level -- it requires the model to integrate the substitution compositionally and infer that the meaning of the larger sentence remains intact. The dataset is released with train (4481 samples), val (387 samples), and test (510 samples) splits.

\textbf{Systematicity: } It refers to a model’s ability to recombine known parts and rules to interpret novel expressions \cite{fodor1988connectionism, hupkes2020compositionality}.  We evaluate this using the \textbf{PLANE Dataset} \cite{bertolini2022testing}, which focuses on adjective–noun (AN) entailment patterns. In this task, the model is provided with two base entailments: one between an adjective–noun phrase and the noun alone (e.g., \textit{red car} $\models$ \textit{car}), and one between the noun and its hypernym (\textit{car} $\models$ \textit{vehicle}). The model is then asked to determine the validity of the composed entailment (\textit{red car} $\models$ \textit{red vehicle}). The dataset is released with train (60012 samples) and test (2016 samples) splits.

To succeed, the model must apply \textit{systematic compositional reasoning}: it must combine the entailment patterns it has already observed to infer the new one. Crucially, this requires understanding how the adjective interacts with both the noun and the hypernym class -- particularly when adjectives vary in type (e.g., intersective vs subsective vs intensional). The task thus tests whether the model can generalize meaning compositionally, rather than relying on shallow heuristics or memorized patterns.

\textbf{Over-generalization}: It refers to a model’s tendency to incorrectly apply compositional rules to expressions that are, in fact, non-compositional \cite{hupkes2020compositionality}. A robust compositional system should be able to distinguish between genuine compositional combinations and compounds that do not follow regular meaning composition. We test this with a novel task using a handcrafted toy dataset of 45 samples in English language---the \textbf{COMPCOMB Dataset}, which we introduce in this work. Each data point consists of a triple: a noun, a compositional adjective–noun (AN) combination involving that noun, and a non-compositional exocentric compound that includes the same noun. For example, in the triple \textit{(coat, trenchcoat, turncoat)}, the word \textit{trenchcoat} refers to a specific type of \textit{coat}, preserving a compositional relationship. In contrast, \textit{turncoat} is an idiomatic expression referring to a traitor, where the meaning does not transparently derive from \textit{coat}.

To succeed, the model must avoid overgeneralizing based on surface forms or string overlap, and instead distinguish between compositional and non-compositional expressions. This task directly tests whether the model can identify and respect the boundaries of compositional generalization. Please note that we will release the dataset upon publication along with the license and intended use for research purposes.

It is important to note that two other dimensions as per \citet{hupkes2020compositionality}---\textit{productivity} and \textit{localism}---are deliberately excluded from our analyses, for reasons rooted in both linguistic theory and the epistemological limitations of empirical verification which are commonly recognized in the literature \cite{hupkes2020compositionality}. 

\textbf{First}, productivity---the notion that natural language allows for the generation of an unbounded number of novel expressions from a finite set of primitives—is a foundational concept in generative linguistics \cite{chomsky1956three}. However, its empirical status is more controversial. As noted by \citet{pullum2010recursion}, it is not possible to establish the infinitude of natural language through observation, given the finiteness of human utterances and cognitive resources. Moreover, even if a system generalizes to longer sequences than it was trained on, this does not, in itself, constitute evidence for true linguistic productivity. In formal terms, productivity involves generalization across an unbounded domain:
\[
\forall n \in \mathbb{N},\ \exists \ s_n \in \mathcal{L} \text{ such that } \text{length}(s_n) = n,
\]
where $\mathcal{L}$ is the set of grammatical sentences. Demonstrating this for either human language or LLM behavior is inherently intractable. Consequently, productivity lacks the falsifiability and granularity required for robust evaluation in our setting.

\textbf{Second}, we exclude localism as a defining criterion of compositionality. Localism refers to the idea that composition should occur via strictly local operations—i.e., the meaning of a complex expression is derived solely from the meanings of its immediate subparts (its ``children'' in a compositional tree). This notion is aligned with what the relevant linguistic literature \citet{pagin2010compositionality, pagin2010compositionality, pagin2010pure} terms \textit{strong} or \textit{first-level} compositionality. However, this is only one possible interpretation. Under \textit{weak} or \textit{global} compositionality, meanings may be computed from broader structures, potentially referencing distant context or higher-order structures. Crucially, both modes of composition satisfy the formal definition of compositionality:
\[
\llbracket E \rrbracket = f(\llbracket E_1 \rrbracket, \ldots, \llbracket E_n \rrbracket),
\]
where $\llbracket E \rrbracket$ denotes the meaning of expression $E$, and $E_1, \ldots, E_n$ are its constituents. The nature of $f$—whether strictly local or globally informed—is not constrained by the principle of compositionality itself. Empirically, the absence of local compositional operations does not entail non-compositional behavior. A model may validly construct meanings using global strategies and still preserve the structural integrity of compositional mappings. Thus, testing for localism does not test for compositionality per se, but rather for a specific implementation strategy.

For these reasons, we limit our analysis to aspects of compositionality that are both theoretically robust and empirically interpretable. We regard productivity and localism as valuable extensions for future work, but not as necessary conditions for evaluating whether compositionality matters in current large language models.

\section{Model Details}
\label{appendix:b}

In this appendix section, we provide details of the models used in our experiments. For each model family, we selected three models for testing:

\begin{itemize}
    \item Base Model (Base): A minimum parameter model, ranging from 2B to 7B,  that serves as the foundational model version for each family.
    \item Instruction -- Finetuned Model (IFT): The same 7B base model, further fine-tuned with instruction-tuning to enhance task performance.
    \item Scaled Model (Large): A model variant with a higher parameter count, ranging from 9B to 70B, depending on availability within each family. These larger models allow us to investigate how scaling affects compositional behavior.
\end{itemize}

The diversity in models ensures that our analysis captures how both model complexity and tuning approaches impact compositionality. The names and links to each model are in \autoref{tab:model_links}

\begin{table*}[ht]
\centering
\renewcommand{\arraystretch}{1.2}
\setlength{\tabcolsep}{6pt}
\resizebox{1\textwidth}{!}{%
\begin{tabular*}{\textwidth}{@{\extracolsep{\fill}}llll}
\toprule
\textbf{Family} & \textbf{Variant} & \textbf{Hugging Face Link} & \textbf{License} \\ 
\midrule

\multirow{3}{*}{LLaMA-2}
& Base (7B) & \href{https://huggingface.co/meta-llama/Llama-2-7b-hf}{Llama-2-7b-hf} & LLaMA 2 Community License \\
& Instruction-tuned (7B) & \href{https://huggingface.co/meta-llama/Llama-2-7b-chat-hf}{Llama-2-7b-chat-hf} & LLaMA 2 Community License \\
& Scaled (13B) & \href{https://huggingface.co/meta-llama/Llama-2-13b-hf}{Llama-2-13b-hf} & LLaMA 2 Community License \\

\midrule

\multirow{3}{*}{CodeLlama}
& Base (7B) & \href{https://huggingface.co/codellama/CodeLlama-7b-hf}{CodeLlama-7b-hf} & LLaMA 2 Community License \\
& Instruction-tuned (7B) & \href{https://huggingface.co/codellama/CodeLlama-7b-Instruct-hf}{CodeLlama-7b-Instruct-hf} & LLaMA 2 Community License \\
& Scaled (13B) & \href{https://huggingface.co/codellama/CodeLlama-13b-hf}{CodeLlama-13b-hf} & LLaMA 2 Community License \\

\midrule

\multirow{3}{*}{Qwen2.5-Coder} 
& Base (7B) & \href{https://huggingface.co/Qwen/Qwen2.5-Coder-7B}{Qwen2.5-Coder-7B} & Apache License 2.0 \\
& Instruction-tuned (7B) & \href{https://huggingface.co/Qwen/Qwen2.5-Coder-7B-Instruct}{Qwen2.5-Coder-7B-Instruct} & Apache License 2.0 \\
& Scaled (14B) & \href{https://huggingface.co/Qwen/Qwen2.5-Coder-14B}{Qwen2.5-Coder-14B} & Apache License 2.0 \\

\midrule

\multirow{3}{*}{Gemma2} 
& Base (2B) & \href{https://huggingface.co/google/gemma-2-2b}{Gemma-2-2B} & Gemma Terms of Use \\
& Instruction-tuned (2B) & \href{https://huggingface.co/google/gemma-2-2b-it}{Gemma-2-2B-it} & Gemma Terms of Use \\
& Scaled (9B) & \href{https://huggingface.co/google/gemma-2-9b}{Gemma-2-9B} & Gemma Terms of Use \\

\bottomrule
\end{tabular*}
}
\caption{Models used in \autoref{sec:expsetup}, grouped by family and variant, with links to their Hugging Face repositories and associated licenses.}
\label{tab:model_links}
\end{table*}

\section{Methodology Details}
\label{appendix:c}

This appendix provides formal details for the two evaluation setups described in Sections~\ref{sec:compab} and~\ref{sec:compknow}.

\subsection{Functional Evaluation Details}
\label{appendix:c1}

\paragraph{AddOne and PLANE.}
For binary entailment tasks, we use two evaluation variants.

\textbf{(1) Generation-based MCQ setup.}
Given an input prompt $x$ and two candidate options (entailment and non-entailment), the model generates an output sequence $\hat{y}$. 
A prediction is considered correct if the generated output contains the correct option label. 
Accuracy is computed as:

\begin{equation}
\text{Acc} = \frac{1}{N} \sum_{i=1}^{N} \mathbb{1}[\hat{y}_i \text{ contains } y_i],
\end{equation}

where $N$ is the number of test examples and $y_i$ is the gold label.

\textbf{(2) Comparative log-probability setup.}
Given the same prompt $x$ and two candidate completions $y^{(1)}$ (entailment) and $y^{(0)}$ (non-entailment), we compute conditional log-probabilities:

\begin{equation}
\log P(y^{(j)} \mid x) 
= \sum_{t=1}^{T_j} \log P(y^{(j)}_t \mid x, y^{(j)}_{<t}),
\end{equation}

where $T_j$ is the token length of completion $y^{(j)}$. 
The predicted label is:

\begin{equation}
\hat{y} = \arg\max_{j \in \{0,1\}} \log P(y^{(j)} \mid x).
\end{equation}

Accuracy is computed as the proportion of correct predictions over the test set.

\paragraph{COMPCOMB.}
For COMPCOMB, we use a prompting setup in which the model is asked to judge which expression is semantically closer to the noun $n$: the compositional adjective--noun phrase $an$ or the exocentric compound $c$. 

We evaluate three prompt variants to mitigate prompt sensitivity. 
For each variant $p_j$, the model outputs a choice $\hat{z}^{(j)} \in \{an, c\}$. 
Accuracy per variant is defined as:

\begin{equation}
\text{Acc}(p_j) = \frac{1}{N} \sum_{i=1}^{N} \mathbb{1}[\hat{z}^{(j)}_i = an],
\end{equation}

where $N$ is the number of triples.

Final reported accuracy is averaged across prompt variants:

\begin{equation}
\text{Acc} = \frac{1}{m} \sum_{j=1}^{m} \text{Acc}(p_j),
\end{equation}

where $m$ denotes the number of prompt templates. 
A prediction is considered correct if the model judges $an$ to be closer to $n$ than $c$.

\subsection{Representational Evaluation Details}
\label{appendix:c2}

\paragraph{Layer-wise Extraction.}
For each model, we extract hidden representations corresponding to the task input from every fifth transformer layer. 
Let $\phi_\ell(x)$ denote the representation of input $x$ at layer $\ell$.

\paragraph{AddOne and PLANE.}
For each layer $\ell$, we train a linear classifier on $\phi_\ell(x)$ to predict the binary entailment label. 
Let $\hat{y}_\ell$ denote the predicted label at layer $\ell$. 
Layer-wise accuracy is computed as:

\begin{equation}
\text{Acc}_\ell = \frac{1}{N} \sum_{i=1}^{N} \mathbb{1}[\hat{y}_{\ell,i} = y_i],
\end{equation}

where $N$ is the number of test examples and $y_i$ is the gold label.

\paragraph{COMPCOMB.}
For each triple $(n, an, c)$ and layer $\ell$, we extract representations 
$\phi_\ell(n)$, $\phi_\ell(an)$, and $\phi_\ell(c)$. 
We compute cosine similarity:

\begin{equation}
\text{sim}_\ell(x,y) = 
\frac{\phi_\ell(x) \cdot \phi_\ell(y)}
{\|\phi_\ell(x)\| \|\phi_\ell(y)\|}.
\end{equation}

A prediction is considered correct if:

\begin{equation}
\text{sim}_\ell(n, an) > \text{sim}_\ell(n, c).
\end{equation}

The aim here is to examine how well the model's internal representations encode the relationships between related and unrelated adjective-noun and exocentric compound-noun pairs. Layer-wise accuracy is defined as the proportion of triples satisfying this inequality. This gives insights into how well the model captures semantic relationships and distinctions between input items (such as distinguishing between a noun and its related and unrelated compounds). This setup allows for an analysis of the model's ability to differentiate semantically related pairs from unrelated ones based on internal representation quality.

\end{document}